\documentclass[letterpaper, 10 pt, conference]{ieeeconf}  % Comment this line out if you need a4paper

\IEEEoverridecommandlockouts                              % This command is only needed if 
\usepackage{graphicx}
\usepackage{amsmath}
\usepackage{amssymb}
\usepackage{subcaption}
\usepackage{color,soul}

\title{MODNet: Motion and Appearance based Moving\\ Object Detection Network for Autonomous Driving}

\author{Mennatullah Siam, Heba Mahgoub, Mohamed Zahran,\\ Senthil Yogamani, Martin Jagersand\\
{\tt\small mennatul@ualberta.ca}, {\tt\small h.mahgoub@fci-cu.edu.eg},\\
{\tt\small senthil.yogamani@valeo.com},
{\tt\small jag@cs.ualberta.ca}
}

\begin{document}

\maketitle
\thispagestyle{empty}
\pagestyle{empty}

\begin{abstract}
For autonomous driving, moving objects like vehicles and pedestrians are of critical importance as they primarily influence the maneuvering and braking of the car. Typically, they are detected by motion segmentation of dense optical flow augmented by a CNN based object detector for capturing semantics. In this paper, our aim is to jointly model motion and appearance cues in a single convolutional network. We propose a novel two-stream architecture for joint learning of object detection and motion segmentation. We designed three different flavors of our network to establish systematic comparison. It is shown that the joint training of tasks significantly improves accuracy compared to training them independently. Although motion segmentation has relatively fewer data than vehicle detection. The shared fusion encoder benefits from the joint training to learn a generalized representation. We created our own publicly available dataset (KITTI MOD) by extending KITTI object detection to obtain static/moving annotations on the vehicles. We compared against MPNet as a baseline, which is the current state of the art for CNN-based motion detection. It is shown that the proposed two-stream architecture improves the mAP score by 21.5\% in KITTI MOD. We also evaluated our algorithm on the non-automotive DAVIS dataset and obtained accuracy close to the state-of-the-art performance. The proposed network runs at 8 fps on a Titan X GPU using a basic VGG16 encoder.

\end{abstract}

\section{Introduction}
Autonomous driving is a rapidly advancing application area with the progress in deep learning. There are two main paradigms in this area: (1) The mediated perception approach which semantically reasons the scene \cite{geiger20143d}\cite{teichmann2016multinet} and then determines the driving decision based on it. (2) The behavior reflex approach that learns end to end the driving decision \cite{bojarski2016end} \cite{xu2016end}. The behavior reflex methods can benefit from semantic reasoning of the environment. For example, an auxiliary loss on semantic segmentation \cite{xu2016end} can be used with end to end learning. On the other hand, in mediated perception semantic reasoning is a central task, followed by the control decision separately. Semantic reasoning of the scene includes object detection, motion detection, depth estimation, object tracking and others. Motion detection is a challenging problem because of the continuous ego-camera motion along with the motion of independent objects. 

\begin{figure}[t!]
    \includegraphics[width=0.47\textwidth]{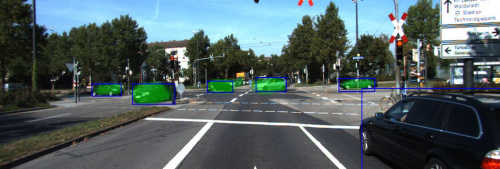}
    \caption{One Forward pass infers vehicle detection and motion segmentation. Green masks represent motion, while blue rectangles represent detected vehicles.}
    \label{fig:overview}
\end{figure}

Moving objects are the most critical in terms of avoiding fatalities and enabling smooth maneuvering and braking of the car. Motion cues can also enable generic object detection as it is not possible to train for all possible object categories beforehand. Classical approaches in motion detection were focused on geometry based approaches \cite{torr1998geometric}\cite{papazoglou2013fast}\cite{ochs2014segmentation}\cite{menze2015object}\cite{scott2017motion}. However, pure geometry based approaches have many limitations, motion parallax issue is one such example. A recent trend  \cite{tokmakov2016learning}\cite{jain2017fusionseg}\cite{drayer2016object}\cite{vijayanarasimhan2017sfm}\cite{fragkiadaki2015learning} for learning motion in videos has emerged. Nonetheless, this trend was focused on pixel-wise motion segmentation.  
Different architectures for the joint reasoning of different tasks were proposed \cite{teichmann2016multinet}\cite{kokkinos2016ubernet}. A shared encoder between these tasks were used, but their work utilizes appearance cues only. 

In this paper, we propose a novel method for scene understanding that combines motion and appearance cues. Scene understanding that relies on appearance cues only, can not infer motion and geometry related information. This includes motion segmentation, optical flow estimation, and depth estimation. In our work we address this gap, and present an example application for joint vehicle detection and motion segmentation, refer to Figure \ref{fig:overview}. The contributions of this work are as follows: (1) We present a novel multi-task learning system for autonomous driving that fuses both appearance and motion cues. (2) This system is used to jointly detect vehicles and segment motion. (3) We propose a method to generate automatically annotated data for this task from KITTI dataset which we call KITTI MOD. This provides a benchmark for autonomous driving application, unlike synthetic sequences \cite{mayer2016large}.

The rest of the paper is organized as follows: Section \ref{sec:related} reviews the related work. Section \ref{sec:method} details the proposed method for incorporating motion cues in motion segmentation and object detection. Section \ref{sec:exps} shows the experimental results and discussion. Finally, section \ref{sec:conc} provides concluding remarks.

\section{Related Work}
\label{sec:related}
\textbf{Object Detection} has seen a lot of progress recently. Mainly two categories have emerged in object detectors. These are region proposals based detectors, and single shot detectors. R-CNN \cite{girshick2016region}, Fast R-CNN \cite{girshick2015fast} and Faster R-CNN \cite{ren2015faster} are examples on the first category. Girshick et. al. proposed R-CNN and Fast R-CNN \cite{girshick2016region}\cite{girshick2015fast} that rely on a separate region proposal module, followed by the detection network. He also proposed a region proposal network incorporated within the detection network in Faster R-CNN \cite{ren2015faster}. 

The second category single shot detectors, do not require a separate proposal generation method. Redmon et. al. \cite{redmon2016you}\cite{redmon2016yolo9000} and Liu et. al. \cite{liu2016ssd} proposed both Yolo and SSD that fall in this category. The Yolo \cite{redmon2016you} method represents the image as a grid of cells. If the center of an object lies in a cell, it would be responsible to estimate that object. Thus, each cell regresses on the box coordinates and size. Each cell estimates the confidence score representing objectness, and class probabilities as well. The continuation of the work in \cite{redmon2016yolo9000} provides a more computationally efficient method, and better average precision. This is mainly due to their use of anchors inspired from Faster R-CNN work, and introducing skip connections for higher resolution feature maps. Single shot detection methods generally provide a more computationally efficient method than generating proposals.

\textbf{Motion Estimation:} Menze et. al. introduced a geometry based approach to estimate scene flow and object motion masks \cite{menze2015object}. However, the approach is computationally expensive with running time 50 minutes per frame. This makes it impractical for autonomous driving. Scott et. al. proposed another geometry based work that models the background motion in terms of a homography \cite{scott2017motion}. It is based on limited assumptions about the camera motion model to include only rotations. This incurred failures with camera translation, which makes it impractical in autonomous driving scenes. Fragiadaki et. al. suggested a method to segment moving objects \cite{fragkiadaki2015learning} that uses a separate proposal generation. This is followed by a moving objectness detector. However, it was previously shown in object detection literature that proposal generation methods are computationally inefficient. Jain et. al. presented a method for appearance and motion fusion in \cite{jain2017fusionseg}. The work focuses on generic object segmentation. It was not designed for static/moving objects classification. 

Tokmakov et. al. \cite{tokmakov2016learning} used a one-stream fully convolutional network with optical flow input to estimate the motion type. The approach works with either optical flow only or concatenated image and flow as input. The concatenated input will not benefit from the available pretrained weights, as they were trained on RGB only. Drayer et. al. \cite{drayer2016object} described a video segmentation work that used tracked detections from R-CNN denoted as tubes. This was followed by a spatiotemporal graph to segment objects. The main issue with this approach is its running time of 8 seconds per frame. Thus, there is a need for an efficient and more accurate solution. 

\section{Method}
\label{sec:method}
In this section both motion and object detection networks are detailed. First, a description of the method for generating motion relevant annotations on KITTI is presented. Then a two-stream architecture to segment pixel-wise motion masks is described. Finally, a method for jointly detecting vehicles and segmenting motion is described. 

\subsection{KITTI MOD Dataset}
Training convolutional networks requires large amounts of training data. We suggest a pipeline to automatically generate static/moving classification for objects. The procedure uses odometry information and annotated 3D bounding boxes for vehicles. The odometry information that includes GPS/IMU readings provides a method to compute the velocity of the moving camera. The 3D bounding boxes of the annotated vehicles are projected to 2D images and tagged with their corresponding 3D centroid. The 2D bounding boxes are associated between consecutive frames using intersection over union. The estimated vehicles velocities are then computed based on the associated 3D centroids. The computed velocity vector per bounding box is compared to the odometry ground-truth to determine the static/moving classification of vehicles. The objects that are then consistently identified on multiple frames as moving are kept. In this dataset, the focus is on vehicles with car, truck, and van object categories. 
\begin{figure}[h!]
    \includegraphics[width=0.5\textwidth]{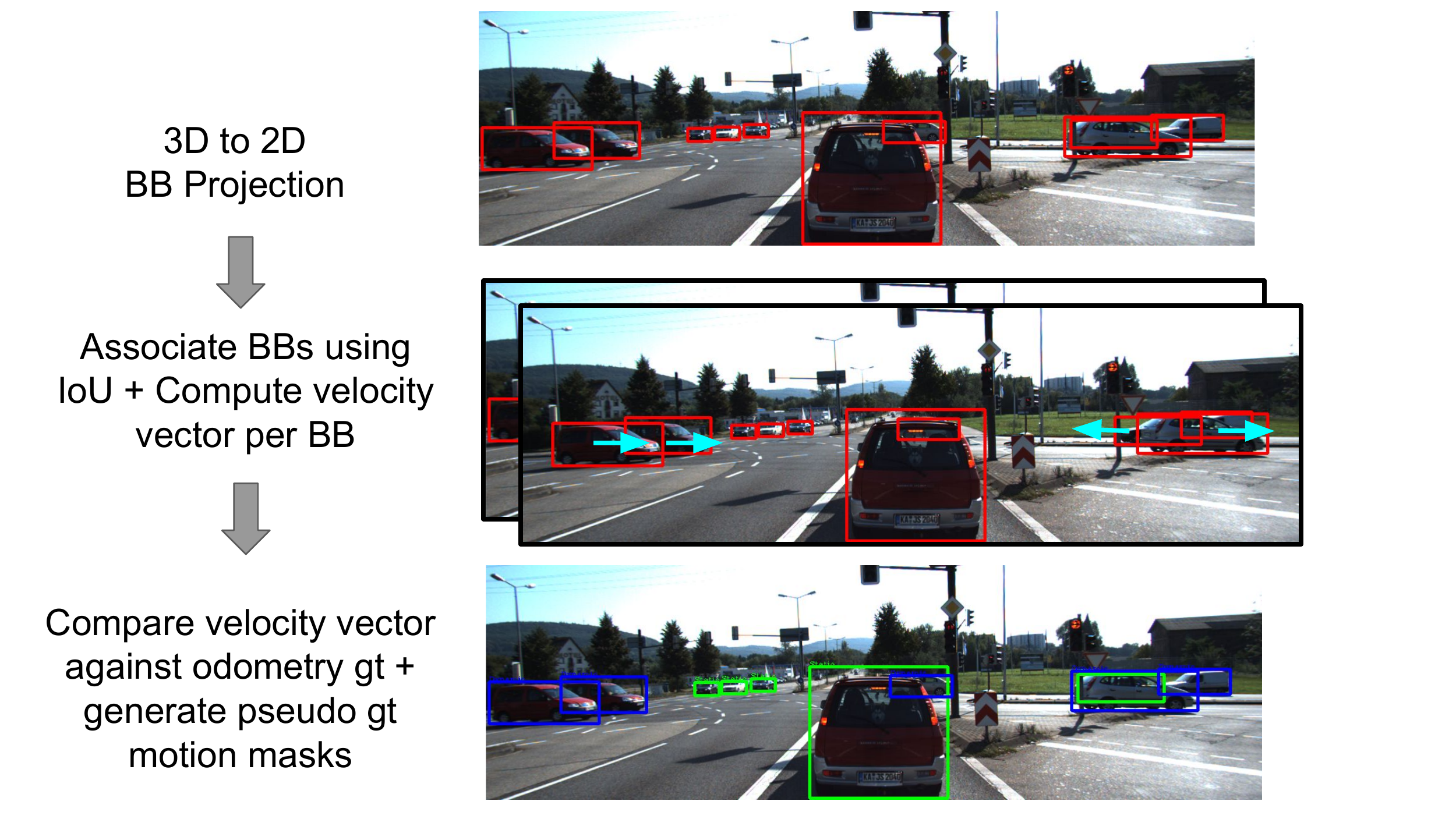}
    \caption{Overview of the pipeline used to generate KITTI Moving Object Detection annotations. Blue boxes for moving vehicles, green boxes for static ones.}
    \label{fig:kittimod}
\end{figure}

An overview of the labeling procedure is shown in Figure \ref{fig:kittimod}. This is applied on six sequences from KITTI raw data \cite{Geiger2013IJRR} to generate a total of 1750 frames. In addition to these frames, 200 frames from KITTI scene flow are used to provide us with 1950 frames in total. This new dataset is referred to as KITTI MOD throughout the paper. For some statistics on the dataset, the total number of static vehicles is 5997, while the number of moving ones is 2383. The dataset is publicly available \cite{multimenna} to act as a benchmark on motion detection on KITTI. Although there exists other motion segmentation datasets such as \cite{perazzi2016}\cite{mayer2016large}\cite{ochs2014segmentation}. However, they are either synthetic\cite{mayer2016large}, relatively small \cite{ochs2014segmentation} or has limited camera motion \cite{perazzi2016} unlike what is present in autonomous driving scenes.
\begin{figure*}[ht!]
    \centering
    \includegraphics[width=0.9\textwidth]{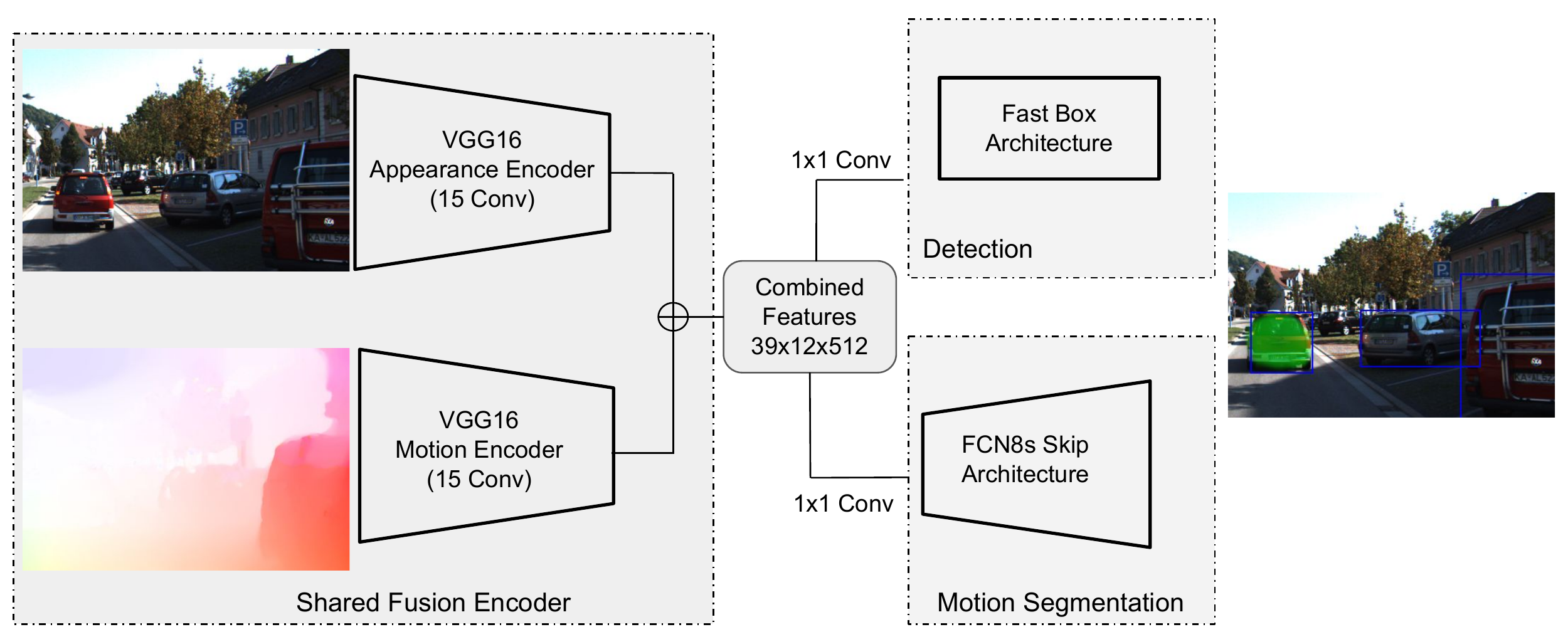}
    \caption{MODNet Two Stream Multi-Task Learning Architecture for joint motion segmentation and object detection. Optical Flow and RGB input, RGB image with overlay motion segmentation in green and detected bounding boxes in blue. Fast Box module regresses on x,y,w,h of each bounding box and confidence score.}
    \label{fig:2st}
\end{figure*}

\subsection{Motion Segmentation}
An encoder-decoder architecture is used for motion segmentation. Similar to the FCN8s \cite{long2015fully} architecture, VGG16 network is transformed to a fully convolutional network. The first 15 convolutional layers are used for feature extraction. However, unlike other segmentation architectures, our network combines motion and appearance cues. Inspiring from \cite{simonyan2014two}\cite{jain2017fusionseg}, a two stream VGG16 is utilized to extract appearance and motion features. The feature maps from both are combined using a summation junction for a memory efficient network. This is followed by 1x1 convolutional layer, then three transposed convolutional layers to perform upsampling. In order to benefit from high-resolution features, skip connections are used and added to partially upsampled feature maps. The appearance stream helps in segmenting the vehicle boundary, while the motion stream identifies moving vehicles.

Two different inputs for the motion stream are considered and compared: (1) Optical flow. (2)  Image pair of frame $I_t, I_{t-1}$. In the latter case, the network is expected to learn an embedding that matches the input image pair. In case of optical flow, a color wheel representation is used to convert it to RGB flow\cite{baker2011database}. The benefit of using such representation is to utilize pretrained VGG16 weights on ImageNet. This helps the network to learn a better-generalized representation instead of training from scratch. Pixel-wise cross entropy loss is used for the segmentation network.

\subsection{Joint Vehicle and Motion Detection}
In autonomous driving, static/moving classification on the object-level is more relevant than dense pixel-level classification. A method that jointly detects vehicles in the scene while classifying them into static/moving is presented. Two approaches are further studied for this purpose. One is to separate the tasks of detection and motion segmentation. The other is to share the two-stream encoder and jointly train for the two tasks. In the first approach, the same two stream architecture is utilized to generate motion masks. A detector similar to the detection decoder in \cite{teichmann2016multinet} denoted as FastBox is used. It is based on Yolo \cite{redmon2016you} used as a single shot detector utilizing the first 15 convolutional layers from VGG16. This is followed by two 1x1 convolutional layers. The last layer outputs 39x12 grid size representing each cell. The channels in the output layer include the bounding box coordinates, size, and the confidence in the existence of a vehicle. Finally, the rezoom layer is used to overcome the loss of resolution caused by pooling. ROI pooling from the higher resolution layers is followed by 1x1 convolutional layers. Then the residuals on the coordinates are regressed over, for a more accurate localization. The loss function used in detection combines the L1 loss for the bounding box regression, with cross entropy for the confidence score. 

In the second approach a shared two-stream VGG16 encoder is used to output the combined motion and appearance features. This is followed by two decoders for vehicle detection and motion segmentation. This network is referred to as moving object detection network (MODNet). This method follows a similar approach to the work in \cite{teichmann2016multinet}. However, in our approach we present motion cues as another valuable input to any multi-task learning network in auto-driving. It also has similarities to the work in \cite{jain2017fusionseg}, but their work did not include joint detection. This is one of the main strengths of our work; In the same forward pass motion segmentation and vehicle detection are predicted. This is crucial for real-time performance in autonomous driving scenarios. Inside the segmentation network for each skip connection a summation junction is used to combine motion and appearance features. The detection decoder utilizes the appearance features only and ignores motion features.
\begin{subequations}
\begin{equation}
L_{total}= L_{seg} + L_{det}
\end{equation}
\begin{equation}
L_{seg}= - \frac{1}{|I|} \sum_{i \in I} \sum_{c \in C_{motion}} p_i(c)\log{q_i(c)}
\end{equation}
\begin{align}
L_{det}= \frac{1}{|S|} \sum_{s \in S} 1^{obj} (|x_{q_s}-x_{p_s}| + |y_{q_s}-y_{p_s}| \nonumber \\
+ |w_{q_s}-w_{p_s}| + |h_{q_s}-h_{p_s}|) \nonumber \\
- \frac{1}{|S|} \sum_{s \in S} \sum_{c' \in C_{vehicle}} p_s(c')\log{q_s(c')}
\end{align}
\label{eq:losses}
\end{subequations}

The loss function alternates between segmentation and detection losses as shown in Equation \ref{eq:losses}. In these equations, $q$ denotes predictions and $p$ denotes ground-truth. The pixel locations are termed as $I$, while $S$ is the grid cells. $C_{motion}$ is the set of classes for motion segmentation as foreground or background, while $C_{vehicle}$ is the classes for vehicle classification. The detection loss regresses with the L1 loss on the coordinates within the cell. Only cells with a positive confidence score are considered in the regression loss. Joint training is performed similarly to \cite{teichmann2016multinet} where gradients are back-propagated from both tasks on their corresponding mini-batch inputs. This method of joint training leverages the performance of tasks with comparably fewer data. This provides another motivation for the shared motion and appearance encoders. It is worth noting, that motion relevant annotations such as motion masks or optical flow groundtruth are relatively small in real datasets. The tasks for training are selected in an alternate fashion with equal probabilities. Finally, a similar network with joint training of motion segmentation, vehicle detection and road segmentation is used. Thus it is able to infer the semantics of the scene in one forward pass.

\section{Experiments}
\label{sec:exps}
In this section, we present the datasets used, experimental setup, and results on both motion segmentation and joint detection and segmentation. 

\subsection{Datasets}
The proposed framework is tested on the challenging KITTI dataset \cite{Geiger2013IJRR}. KITTI scene flow \cite{menze2015object}, our generated KITTI MOD data and The Davis \cite{perazzi2016} benchmark are used. DAVIS is comprised of 50 sequences, with 3455 total number of frames. However, it does not include fast forward moving camera unlike KITTI sequences. The objects moving along the same camera direction, poses another challenge not available in DAVIS. Most of the sequences are dominated by two or three salient objects in the whole scene. Motion segmentation is initially evaluated on KITTI Scene Flow data and then DAVIS. Then the moving object detection network (MODNet) is trained and evaluated on KITTI MOD dataset.

\subsection{Experimental Setup}
Throughout experiments, the Adam optimizer is used with learning rate $1e^{-5}$. L2 regularization is used in the loss function to avoid overfitting the data, with $5e^{-4}$ factor. Dropout with probability 0.5 is used to 1x1 convolutional layers. The encoder is initialized with VGG pretrained weights on Imagenet. Transposed convolution layers are initialized to bilinear upsampling. Input image resolution used is 1048x384. 

The evaluation metrics used in segmentation are precision, recall, F-score and mean intersection over union (IoU). The evaluation metric used for detection is mean average precision(mAP) and average precision (AP) for static/moving classes. Average precision of car class is also measured showing different difficulties for easy, medium, and hard setup as in KITTI benchmark \cite{Geiger2012CVPR}. Note that it is important to evaluate the static/moving classification standalone without including errors from the detection itself. The average precision used is computed on the detected bounding boxes that match bounding boxes from the ground truth. Thus, evaluation is for static/moving classification standalone, without penalizing errors from FastBox detection.

\subsection{Experimental Results}
\subsubsection{Motion Segmentation on KITTI}
 Initial experiments for motion segmentation on KITTI sceneflow is conducted. The goal is to initially compare image pair against optical flow representation as input. These results are shown in Table \ref{table:quant_seg}. It compares the quantitative evaluation of our two-stream motion segmentation network against the one stream optical flow. The two-stream (RGB+OF) shows a 10\% increase in average IoU over the one-stream, since the appearance stream pushes toward better vehicle boundary segmentation. The two-stream architecture with image and optical flow as input(RGB+OF) and with image pair input is compared.  The image-pair method struggles more than (RGB + OF). This is expected as optical flow input is a better motion representation to the network.

\begin{table}[ht!]
\centering
\caption{Quantitative evaluation on KITTI data for our proposed two-stream motion segmentation network. }
\begin{tabular}{|l|l|l|l|l|}
\hline
 & Precision & Recall & F-Score & IoU \\ \hline
1 Stream & 70.4 & 45.66 & 38.31 & 50.4\\ \hline
2 Stream (image pair) & 76.4 & 67.68 & 71.78 & 55.98\\ \hline
2 Stream (RGB+OF) & \textbf{74.07} & \textbf{76.38} & \textbf{75.2} & \textbf{60.27} \\ \hline
\end{tabular}
\label{table:quant_seg}
\end{table}

\subsubsection{Joint Motion Segmentation and Vehicle Detection} Detailed experiments on motion segmentation with vehicle detection is conducted on KITTI MOD. Table \ref{table:quant_segkittimod} shows the evaluation of the separate and joint training for motion segmentation and vehicle detection. The detection evaluation for the separate setup is taken from \cite{teichmann2016multinet} since their pre-trained weights are used in this setup. It clearly shows that the joint training improves the motion segmentation with 8.2\% approximately in F-score. The detection on the easy evaluation is only affected by 2.5\% and on the hard evaluation is approximately the same. It is worth noting that joint training of both tasks improves results when there is limited training data.
%\begin{figure}[ht!]
%\begin{subfigure}{0.5\textwidth}
%    \includegraphics[scale= 0.45]{Images/sep-train}
%\end{subfigure}

%\begin{subfigure}{1.0\textwidth}
%    \includegraphics[scale= 1.0]{Images/joint-train}
%\end{subfigure}
%\end{figure}

\begin{table*}[ht!]
\centering
\caption{Quantitative comparison on KITTI MOD data for separate MODNet against jointly trained MODNet.}
\begin{tabular}{|l|l|l|l|l|l|l|l|}
\hline
 & \multicolumn{3}{|c|}{Object Detection} & \multicolumn{4}{|c|}{Motion Segmentation} \\\hline
  & moderate & easy & hard & Precision & Recall & F-score & IoU \\ \hline
MODNet (RGB+OF)- Separate & \textbf{83.35} & \textbf{92.8} & 67.59 & 44.34 & 69.84 & 54.25 & 37.22\\ \hline
MODNet (RGB+OF)- Joint & 80.74 &  89.52 & \textbf{67.72} & \textbf{56.18} & \textbf{70.32} & \textbf{62.46} & \textbf{45.41}\\ \hline
\end{tabular}
\label{table:quant_segkittimod}
\end{table*}

\begin{figure*}[ht!]
\begin{subfigure}{.5\textwidth}
    \includegraphics[scale= 0.55]{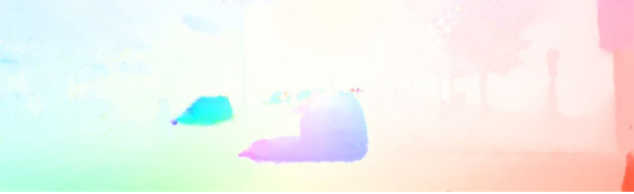}
\end{subfigure}%
\begin{subfigure}{.5\textwidth}
    \includegraphics[scale= 0.55]{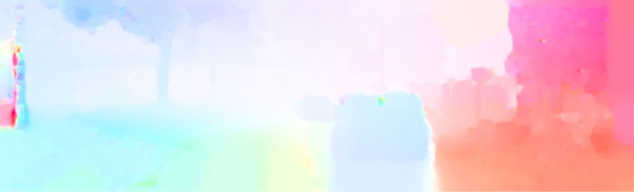}
\end{subfigure}

\begin{subfigure}{.5\textwidth}
    \includegraphics[scale= 0.55]{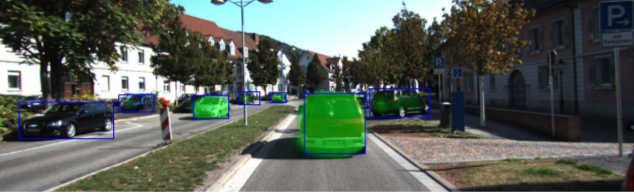}
\end{subfigure}%
\begin{subfigure}{.5\textwidth}
    \includegraphics[scale= 0.55]{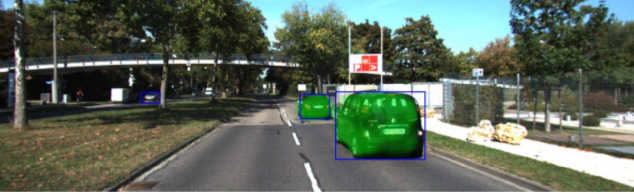}
\end{subfigure}

\begin{subfigure}{.5\textwidth}
    \includegraphics[scale= 0.55]{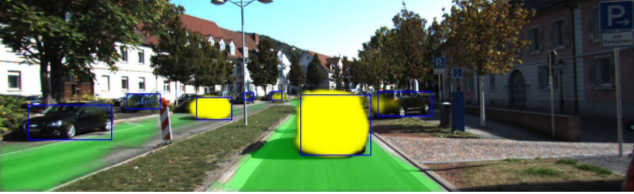}
\end{subfigure}%
\begin{subfigure}{.5\textwidth}
    \includegraphics[scale= 0.55]{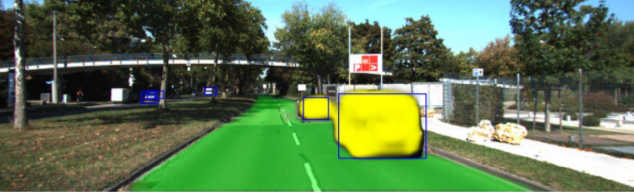}
\end{subfigure}
    \caption{Qualitative evaluation on KITTI MOD data for our proposed two-stream multi-task learning network MODNet. top row: Input Optical Flow, middle row output of 2 tasks: overlay motion mask (green), bottom row output of 3 tasks: overlay motion mask (yellow), road segmentation (green) and detected bounding boxes (blue).}
    \label{fig:qual_seg}
\end{figure*}

The two-stream motion segmentation network is used to provide motion masks which are then combined with FastBox \cite{teichmann2016multinet} detections. The output segmentation and vehicles' static/moving classification is evaluated on KITTI MOD data. Table \ref{table:quant_det} shows the results from the joint detection and motion segmentation. The two-stream MODNet shows the best mAP on KITTI MOD data. This is compared against one of the state-of-the-art methods MPNet \cite{tokmakov2016learning}. MPNet with optical flow input is evaluated on KITTI MOD and combined with proposals as mentioned in their method. Its pretrained weights are used as is, then its output motion segmentation is used with vehicle detection. If intersection over union is larger than 0.5, the detected vehicle is considered moving. This is applied for both our approach and MPNet. It is worth noting that our method for evaluating static/moving classification does not depend on the object detection itself as explained earlier. 

\begin{figure*}[ht!]
\centering
\begin{subfigure}{.48\textwidth}
    \includegraphics[scale= 0.2]{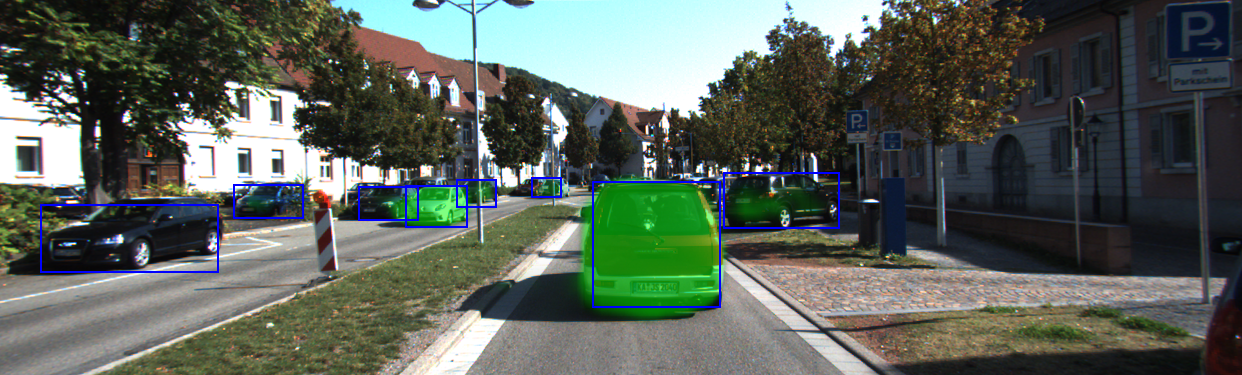}
\end{subfigure}%
\begin{subfigure}{.48\textwidth}
    \includegraphics[scale= 0.2]{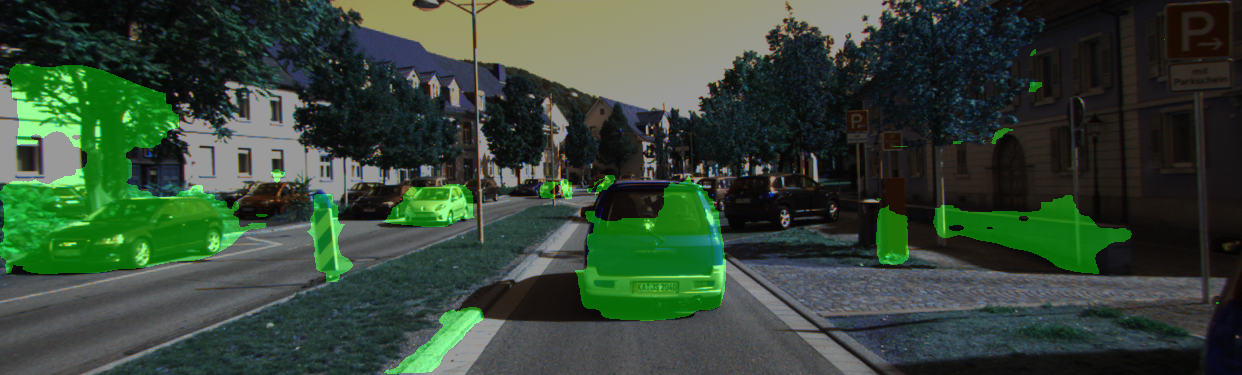}
\end{subfigure}

\begin{subfigure}{.48\textwidth}
    \includegraphics[scale= 0.2]{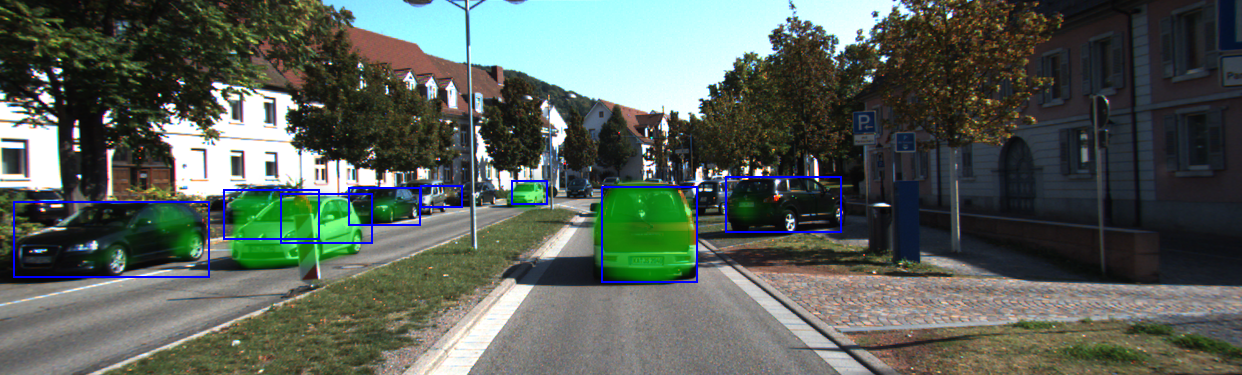}
\end{subfigure}%
\begin{subfigure}{.48\textwidth}
    \includegraphics[scale= 0.2]{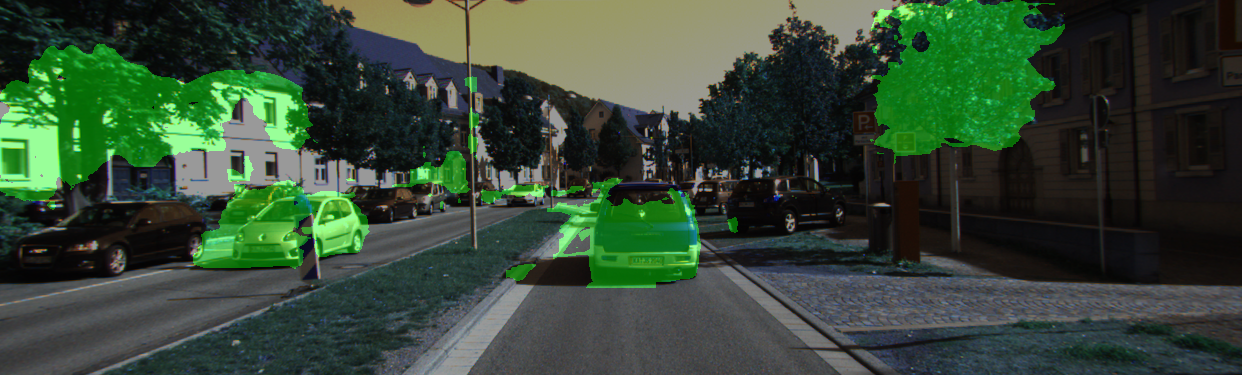}
\end{subfigure}
    \caption{Qualitative comparison on KITTI MOD data for our proposed two-stream multi-task learning network MODNet against MPNet. Green overlay for motion masks.}
    \label{fig:qual_comp}
\end{figure*}

Our proposed approach outperforms MPNet with 21.5\% in mAP. Qualitative comparisons between our proposed work MODNet and MPNet are shown in Figure\ref{fig:qual_comp}. This shows that autonomous driving scenarios, exhibit different challenges compared to generic object segmentation. The continuous camera motion and the existence of multiple objects in the scene makes it to more challenging. The reasons behind our improvement is two fold. The KITTI MOD training data provide a better representation for motion than the synthetic data used in MPNet. The usage of both optical flow and RGB in a two-stream network that utilizes pretrained VGG16 weights improves the results even more. The two-stream image pair is worse in mAP compared to (RGB+OF), but it is more computationally efficient. The joint detection and motion segmentation method provides an efficient way to solve both tasks. Our method runs at 8 fps on a TITANX GPU. This outperforms other approaches in the literature in terms of computational efficiency. The running time for approaches that estimate scene flow can be up to 50 minutes, while the approach in \cite{drayer2016object} takes up to 8 seconds per frame.

\begin{table}[ht!]
\centering
\caption{Quantitative evaluation on KITTI MOD data for our proposed joint detection and motion segmentation network.}
\begin{tabular}{|l|l|l|l|}
\hline
 & AP Static & AP Moving & mAP \\ \hline
MPNet\cite{tokmakov2016learning} & 50.23 & 31.84 & 41.03 \\ \hline
MODNet (image pair) & 60.7 & 44.29 & 52.5\\ \hline
MODNet (RGB+OF)- Separate& \textbf{65.28} & 56.86 & 61.07 \\ \hline
MODNet (RGB+OF)- Joint& 58.6 & \textbf{66.54} & \textbf{62.57} \\ \hline
\end{tabular}
\label{table:quant_det}
\end{table}

\begin{figure*}[ht!]
\centering
\begin{subfigure}{.3\textwidth}
    \includegraphics[scale= 0.2]{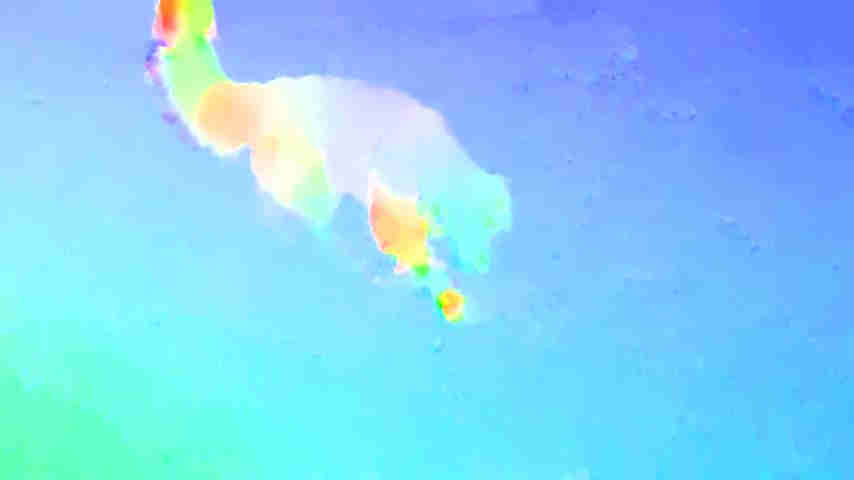}
\end{subfigure}%
\begin{subfigure}{.3\textwidth}
    \includegraphics[scale= 0.2]{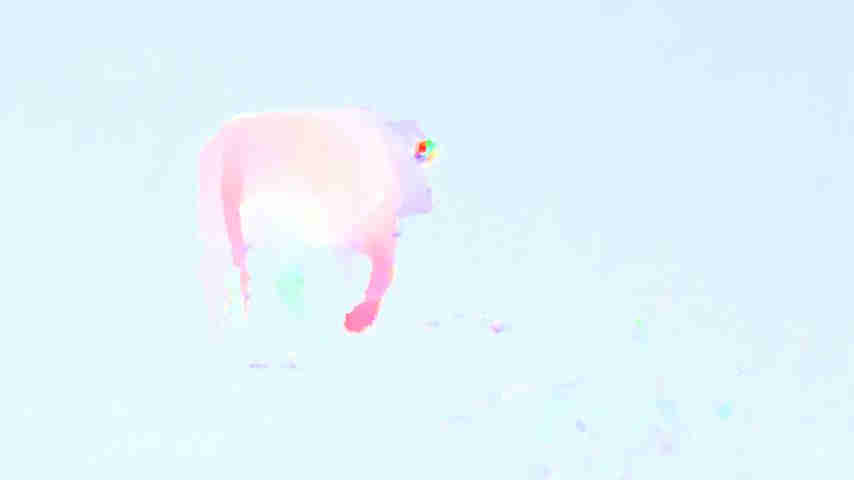}
\end{subfigure}%
\begin{subfigure}{.3\textwidth}
    \includegraphics[scale= 0.2]{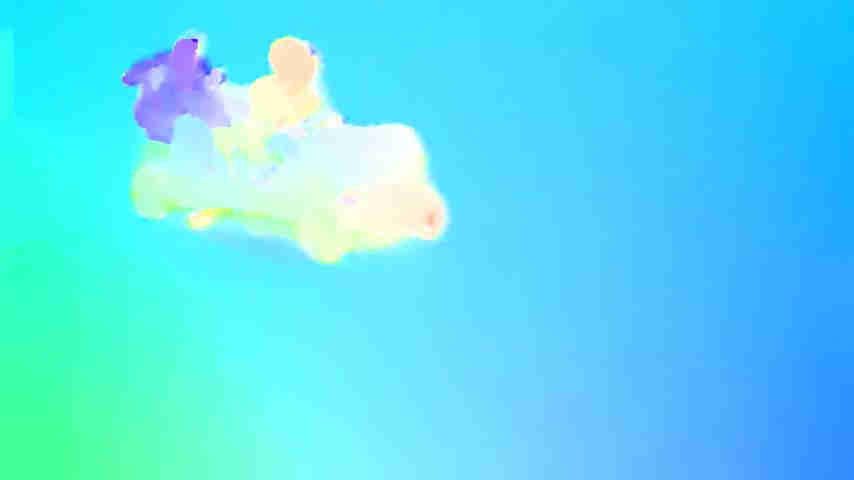}
\end{subfigure}

\begin{subfigure}{.3\textwidth}
    \includegraphics[scale= 0.2]{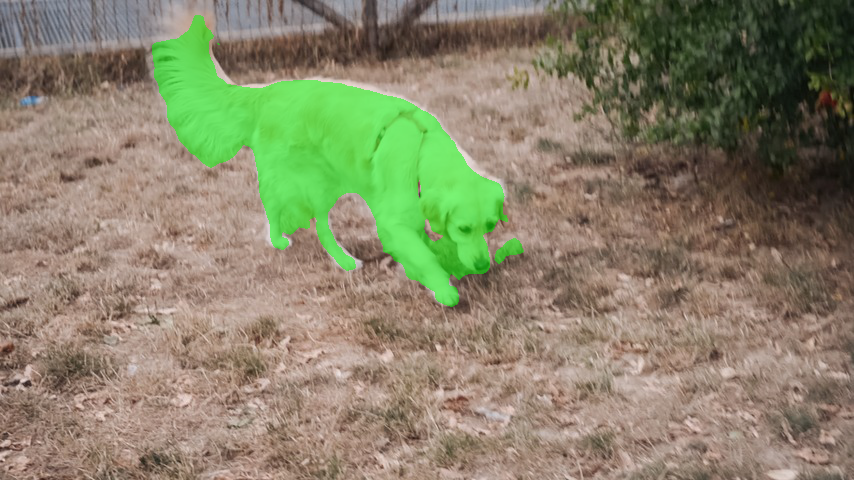}
\end{subfigure}%
\begin{subfigure}{.3\textwidth}
    \includegraphics[scale= 0.2]{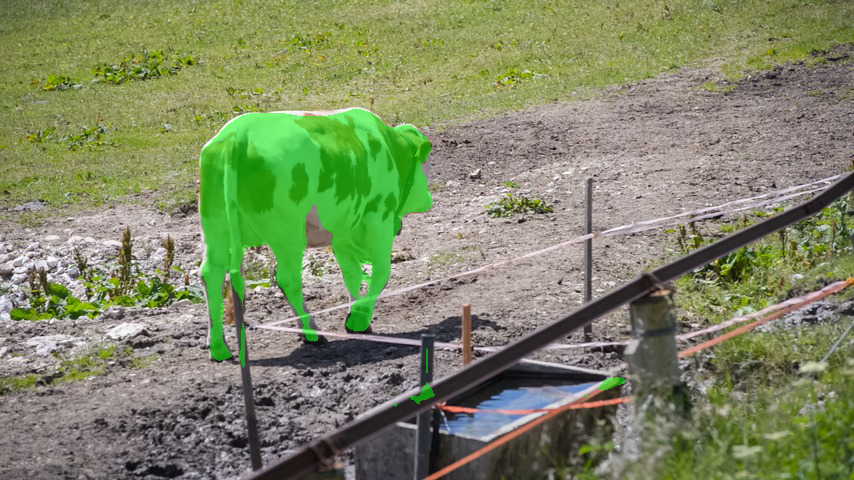}
\end{subfigure}%
\begin{subfigure}{.3\textwidth}
    \includegraphics[scale= 0.2]{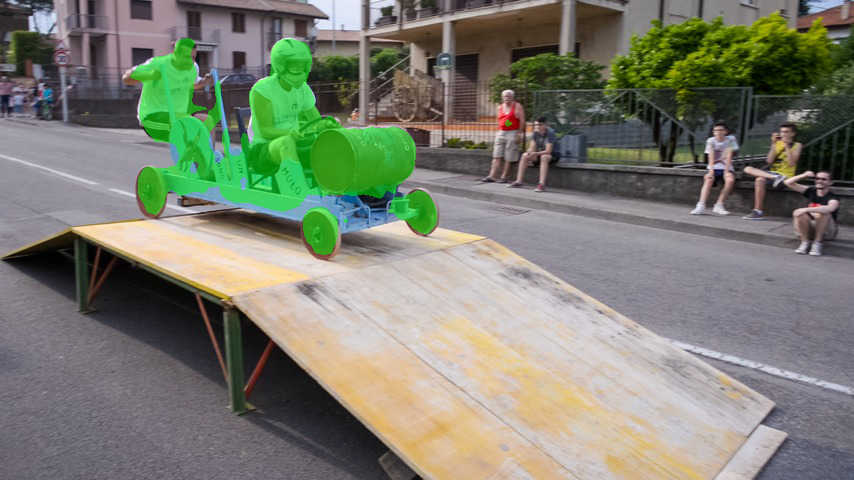}
\end{subfigure}
    \caption{Qualitative evaluation on DAVIS for our proposed two-stream motion segmentation network. RGB Image, Optical Flow and Overlay Motion mask in green.}
    \label{fig:qual_davis}
\end{figure*}

\begin{table*}[ht!]
\centering
\caption{Quantitative evaluation on Davis\cite{perazzi2016} data Val 2016 using mean IoU. Approaches highlighted in blue are without CRF post-processing, and in red after post-processing.}
\label{my-label}
\begin{tabular}{|l|l|l|l|l|l|l|l|l|l|l|}
\hline
 & NLC\cite{faktor2014video} & CVOS\cite{taylor2015causal} & KEY\cite{lee2011key} & MSG\cite{brox2010object} & FST\cite{papazoglou2013fast} & BMM\cite{scott2017motion} & MPNet\cite{tokmakov2016learning} & MPNet\cite{tokmakov2016learning}+CRF & ours & ours+CRF\\ \hline
mIoU & 55.1 & 48.2 & 49.8 & 53.3 & 55.8 & 62.5 & \textcolor{blue}{62.66} & \textcolor{red}{\textbf{70.0}} & \textcolor{blue}{\textbf{63.88}} & \textcolor{red}{66.0} \\ \hline
\end{tabular}
\label{table:quant_davis}
\end{table*}
\subsubsection{Generic Motion Segmentation on DAVIS}
     To additionally compare against the state of the art in segmentation, our method is evaluated on the Davis\cite{perazzi2016} benchmark. MODNet is trained on DAVIS training data and evaluated on the validation set. Then it is compared to the unsupervised methods on DAVIS video segmentation benchmark. Note that on DAVIS the term unsupervised denotes that no masks from the initial frame is used as initialization. MPNet is one of the unsupervised methods that works with one stream only and optical flow as input. It is evaluated with and without applying conditional random fields as a post processing, and with the usage of optical flow only. Table \ref{table:quant_davis} shows that our method outperforms the state of the art on DAVIS in unsupervised motion segmentation, except for MPNet+CRF. The improvement over MPNet alone is only 1.5\%. MPNet+CRF performs better than ours+CRF, but conditional random field runs in 1.15 seconds per frame. This was measured using input image resolution of 480x854 on an Intel core i5 CPU at 2.30 GHZ. Hence, the usage of CRF as postprocessing is impractical for real-time autonomous driving tasks. 

The DAVIS data has very simple camera motion compared to KITTI, so the KITTI MOD dataset poses challenging conditions, different from DAVIS. Another difference between KITTI sequences and DAVIS is that moving objects cover large portions of the scene. Thus, using optical flow can be sufficient for segmentation. Figure \ref{fig:qual_davis} shows the optical flow and segmentation output from our approach on DAVIS data.

\section{Conclusion}
\label{sec:conc}
In this paper, we explore the problem of moving object detection for autonomous driving. We propose a novel two-stream architecture which jointly estimates the motion mask and object detection. Four architectures have been designed and compared: (1) one stream with optical flow, (2) two streams with optical flow and RGB trained separately, (3)  two streams with optical flow and RGB trained jointly and  (4) two streams with consecutive images. Experimental results show that the combined appearance and motion cues in a multi-task learning system outperforms the other architectures. To our knowledge, we are the first to jointly model motion and appearance cues for moving object detection. This provides the flexibility to detect untrained objects purely based on motion cues for rare vehicles like construction trucks. Our approach outperforms the single-stream state-of-the-art MPnet by 21.5\% in mAP on the extended KITTI dataset (KITTI MOD). To conclude, this problem is still far from being solved and deployed in a real-world system and the main bottleneck is the lack of large varied datasets for motion segmentation.

{\small
\bibliographystyle{IEEEtranS}
\bibliography{IEEEfull}
}

\end{document}